\documentclass[10pt,journal,compsoc]{IEEEtran}
\pdfoutput=0
\ifCLASSOPTIONcompsoc
  \usepackage[nocompress]{cite}
\else
  \usepackage{cite}
\fi

\usepackage{times}
\usepackage{graphicx}
\usepackage{amsmath}
\usepackage{amssymb}
\usepackage{booktabs}
\usepackage[pagebackref=true,breaklinks=true,letterpaper=true,colorlinks,bookmarks=false]{hyperref}
\usepackage{url}
\usepackage{multirow}
\usepackage{color}
\usepackage{threeparttable}
\usepackage{balance}

\def \c {{\mathbf{c}}}

\def \y {{\mathbf{l}}}

\def \x {{\mathbf{x}}}
\def \y {{\mathbf{y}}}

\newcommand{\etal}{et al.}

\begin{document}
\title{Face Search at Scale: 80 Million Gallery}

\author{Dayong~Wang,~\IEEEmembership{Member,~IEEE,}
        Charles~Otto,~\IEEEmembership{Student Member,~IEEE,}
        Anil K. Jain,~\IEEEmembership{Fellow,~IEEE}}

\markboth{MSU technical report MSU-CSE-15-11, July 24, 2015}
{Shell \MakeLowercase{\textit{et al.}}: Bare Demo of IEEEtran.cls for Computer Society Journals}

\IEEEtitleabstractindextext{%
\begin{abstract}
Due to the prevalence of social media websites, one challenge facing computer vision researchers is to devise methods to process and search for persons of interest among the billions of shared photos on these websites.
Facebook revealed in a 2013 white paper that its users have uploaded more than $250$ billion photos, and are uploading $350$ million new photos each day. Due to this humongous amount of data, large-scale face search for mining web images is both important and challenging. Despite significant progress in face recognition, searching a large collection of unconstrained face images has not been adequately addressed. To address this challenge, we propose a face search system which combines a fast search procedure, coupled with a state-of-the-art commercial off the shelf (COTS) matcher, in a cascaded framework. Given a probe face, we first filter the large gallery of photos to find the top-$k$ most similar faces using deep features generated from a convolutional neural network. The $k$ retrieved candidates are re-ranked by combining similarities from deep features and the COTS matcher. We evaluate the proposed face search system on a gallery containing $80$ million web-downloaded face images. Experimental results demonstrate that the deep features are competitive with state-of-the-art methods on unconstrained face recognition benchmarks (LFW and IJB-A). More specifically, on the LFW database, we achieve $98.23\%$ accuracy under the standard protocol and a verification rate of $87.65\%$ at FAR of $0.1\%$ under the BLUFR protocol. For the IJB-A benchmark, our accuracies are as follows: TAR of $51.4\%$ at FAR of $0.1\%$ (verification); Rank 1 retrieval of $82.0\%$ (closed-set search); FNIR of $61.7\%$ at FPIR of $1\%$ (open-set search). Further, the proposed face search system offers an excellent trade-off between accuracy and scalability on datasets consisting of millions of images. Additionally, in an experiment involving searching for face images of the Tsarnaev brothers, convicted of the Boston Marathon bombing, the proposed cascade face search system could find the younger brother's (Dzhokhar Tsarnaev) photo at rank $1$ in $1$ second on a $5$M gallery and at rank $8$ in $7$ seconds on an $80$M gallery.
\end{abstract}

}

\maketitle
\IEEEdisplaynontitleabstractindextext
\IEEEpeerreviewmaketitle

\section{Introduction}\label{sec:introduction}
\begin{figure}
  \centering
  \includegraphics[width=3in]{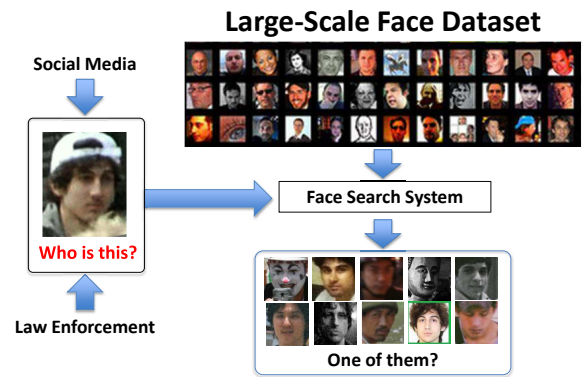}\\
  \caption{An example of large-scale face search problem.}\label{fig:intro}
\end{figure}

Social media has become pervasive in our society. It is hence not surprising that a growing segment of the population has a Facebook, Twitter, Google, or Instagram account. One popular aspect of social media is the sharing of personal photographs. Facebook revealed in a 2013 white paper that its users have uploaded more than 250 billion photos, and are uploading 350 million new photos each day\footnote{\url{https://goo.gl/FmzROn}}. To enable automatic tagging of these images, strong face recognition capabilities are needed. Given an uploaded photo, Facebook and Google's tag suggestion systems automatically detect faces and then suggest possible name tags based on the similarity between facial templates generated from the input photo and previously tagged photographs in their datasets. In the law enforcement domain, the FBI plans to include over $50$ million photographs in its Next Generation Identification (NGI) dataset\footnote{\url{goo.gl/UYlT8p}}, with the goal of providing investigative leads by searching the gallery for images similar to a suspect's photo. Both tag suggestion in social networks and searching for a suspect in criminal investigations are examples of the face search problem (Fig.~\ref{fig:intro}). We address the large-scale face search problem in the context of social media and other web applications where face images are generally unconstrained in terms of pose, expression, and illumination~\cite{faceretrieval:chen2012, faceretrieval:wu2010}.

The major focus in face recognition literature lately has been to improve face recognition accuracy, particularly on the Labeled Faces in the Wild (LFW) dataset~\cite{DB:LFWTech}. But, the problem of scale in face recognition has not been adequately addressed\footnote{An earlier version of the paper appeared in the Proc. IEEE International Conference on Biometrics (ICB), Phuket, June 2015~\cite{icb2015}.}. It is now accepted that the small size of the LFW dataset ($13,233$ images of $5,749$ subjects) and the limitations in the LFW protocol do not address the two major challenges in large-scale face search:
(i) loss in search accuracy with the size of the dataset, and (ii) increase in computational complexity  with dataset size.

The typical approach to scalability (e.g. content-based image retrieval~\cite{faceretrieval:wu2010}) is to represent objects with feature vectors and employ an indexing or approximate search scheme in the feature space. A vast majority of face recognition approaches, irrespective of the representation scheme, are ultimately based on fixed length feature vectors, so employing feature space methods is feasible. However, some techniques are not compatible with feature space approaches such as pairwise comparison models (e.g. Joint-Bayes~\cite{ml:jointbayes}), which have been shown to improve face recognition accuracy.  Additionally, most COTS face recognition SDKs define pairwise comparison scores but do not reveal the underlying feature vectors, so they are also incompatible with feature-space approaches. Therefore, using a feature space based approximation method alone may not be sufficient.

To address the issues of search performance and search time for large datasets ($80$M face images used here), we propose a cascaded face search framework (Fig.~\ref{fig:framework}). In essence, we decompose the search problem into two steps: (i) a fast filtering step, which uses an approximation method to return a short candidate list, and (ii) a re-ranking step, which re-ranks the candidate list with a slower pairwise comparison operation, resulting in a more accurate search. The fast filtering step utilizes a deep convolutional network (ConvNet), which is an efficient implementation of the architecture in~\cite{DB:CASIA}, with product quantization (PQ)~\cite{retrieval:pq}. For the re-ranking step, a COTS face matcher (one of the top performers in the 2014 NIST FRVT~\cite{fvrt:2014}) is used. The main contributions of this paper are as follows:
\begin{itemize}
  \item An efficient deep convolutional network for face recognition, trained on a large public domain data (CASIA~\cite{DB:CASIA}), which improves upon the baseline results reported in~\cite{DB:CASIA}.
  \item A large-scale face search system, leveraging the deep network representation combined with a state-of-the-art COTS face matcher in a cascaded scheme.
  \item Studies on three types of face datasets of increasing complexity: the PCSO mugshot dataset, the LFW dataset (only includes faces detectable by a Viola-Jones face detector), and the IJB-A dataset (includes faces which are not automatically detectable).
  \item The largest face search experiments conducted to date on the LFW~\cite{DB:LFWTech} and IJB-A~\cite{db:janus} face datasets with an $80$M gallery.
  \item Using face images of the Tsarnaev brothers involved in the Boston Marathon bombing as queries, we show that Dzhokhar Tsarnaev's photo could be identified at rank $8$ when searching against the $80$M gallery.
\end{itemize}

The rest of this paper is organized as follows. Section 2 reviews related work on face search. Section 3 details the proposed deep learning architecture and large-scale face search framework. Section 4 introduces the face image datasets used in our experiments. Section 5 presents experiments illustrating the performance of the deep face representation features on face recognition tasks of increasing difficulty (including public domain benchmarks). Section 6 presents large-scale face search results (with $80$M web downloaded face images). Section 7 presents a case study based on the Tsarnev brothers, convicted in the 2013 Boston Marathon bombing. Section 8 concludes the paper.

\begin{table*}[htbp]
\begin{minipage}{\textwidth}
\centering
\caption{A summary of face search systems reported in the literature}\label{tab:fr}
\begin{tabular}{lrrrrll}
\toprule
\multirow{2}{*}{Authors} & \multicolumn{2}{c}{Probe} & \multicolumn{2}{c}{Gallery} & \multirow{2}{*}{Dataset} & \multirow{2}{*}{Search Protocol} \\ \cmidrule(lr){2-3} \cmidrule(lr){4-5}
            & \# Images  & \# Subjects & \# Images & \# Subjects &              &                    \\ \midrule
Wu et al.~\cite{faceretrieval:wu2010}     & $220$    & N/A    & $1$M+ & N/A   & LFW~\cite{DB:LFWTech} + Web Faces\footnote{Web Faces are downloaded from the Internet and used to augment the gallery; different face search systems use their own web faces.} & closed set\\
Chen et al.~\cite{faceretrieval:chen2012} & $120$    & $12$ & $13,113$    & $5,749$ & LFW~\cite{DB:LFWTech} & closed set\\
                                          & $4,300$  & $43$ & $54,497$    & $200$   & Pubfig~\cite{CAVE_0296} & closed set\\
Miller et al.~\cite{miller2015megaface}   & $4,000$  & $80$ & $1$M+ & N/A       & FaceScrub~\cite{db:facecrab} + Yahoo Images\footnote{\url{http://labs.yahoo.com/news/yfcc100m/}} & closed set\\
Yi et al.~\cite{fr:fastmatching}          & $1,195$  & N/A    & $201,196$   & N/A       & FERET~\cite{db:FERET} + Web Faces & closed set \\
Yan et al.~\cite{fr:sh}                   & $16,028$ &  N/A   & $116,028$   & N/A       & FRGC~\cite{db:frgc} + Web Faces & closed set \\
Klare et al.~\cite{fr:cots_cluster}       & $840$    & $840$& $840$       &$840$    & LFW~\cite{DB:LFWTech} & closed set\\
                                          & $25,000$ & $25,000$ & $25,000$ & $25,000$ & PCSO~\cite{fr:cots_cluster} & closed set\\
Best-Rowden et al.~\cite{faceretrieval:fusion} & $10,090$ & $5,153$ & $3,143$ & $596$   & LFW~\cite{DB:LFWTech} & open set \\
Liao et al.~\cite{BLUFR}                       & $8,707$  & $4,249$ & $1,000$ & $1,000$ & LFW~\cite{DB:LFWTech} & open set \\
\midrule
Proposed System & $7,370$ & $5,507$ & $80$M+ & N/A &  LFW~\cite{DB:LFWTech} + Web Faces & closed \& open set\\
                & $5,828$ & $4,500$ & $80$M+ & N/A &  IJB-A~\cite{db:janus} + LFW~\cite{DB:LFWTech} + Web Faces & closed \& open set\\
\bottomrule
\end{tabular}
\end{minipage}
\end{table*}

\section{Related Work}
Face search has been extensively studied in multimedia and computer vision literature~\cite{book:HBforFR2011}. Early studies primarily focused on faces captured under constrained conditions, e.g. the FERET dataset~\cite{db:FERET}. However,
due to the growing need for strong face recognition capability in the social media context, ongoing research is focused on faces captured under more challenging conditions in terms of pose, expression, illumination and aging, similar to images in the public domain datasets LFW~\cite{DB:LFWTech} and IJB-A~\cite{db:janus}.

The three main challenges in large-scale face search are: i) \emph{face representation}, ii) \emph{approximate $k$-NN search}, and iii) \emph{gallery selection and evaluation protocol}. For the face representation, features learned from deep networks (deep features) have been shown to saturate performance on the standard LFW evaluation protocol\footnote{\url{http://vis-www.cs.umass.edu/lfw/results.html}}.
For example, the best recognition performance reported to date on LFW ($99.65\%$)~\cite{dl:tencent} used a deep learning approach leveraging training with $1$M images of $20$K individuals (outside the protocol). A comparable result ($99.63\%$) was achieved by a Google team~\cite{dl:facenet} by training a deep model with about $150$M images of $8$M subjects. It has even been reported that deep features exceed the human face recognition accuracy ($99.20\%$~\cite{CAVE_0296}) on the LFW dataset. To push the frontiers of unconstrained face recognition, the IJB-A dataset was released in 2015~\cite{db:janus}. IJB-A contains face images that are more challenging than LFW in terms of both face detection and face recognition. In order to recognize web downloaded unconstrained face images, we also adopt a deep learning based face representation by improving the architecture outlined in~\cite{DB:CASIA}.

Given our goal of using deep features to filter a large gallery to a small set of candidate face images, we use approximate $k$-NN search to improve scalability. There are three main approaches for approximate face search:
\begin{itemize}
  \item \emph{Inverted Indexing}. Following the traditional bag-of-words representation~\cite{cbir}, Wu et al.~\cite{faceretrieval:wu2010} designed a component-based local face representation for inverted indexing. They first split aligned face images into a set of small blocks around the detected facial landmarks and then quantized each block into a visual word using an identity-based quantization scheme. The candidate images were retrieved from the inverted index of visual words. Chen et al.~\cite{faceretrieval:chen2012} improved the search performance in ~\cite{faceretrieval:wu2010} by leveraging human attributes.
  \item \emph{Hashing}. Yan et al.~\cite{fr:sh} proposed a spectral regression algorithm to project facial features into a discriminative space; a cascaded hashing scheme (similarity hashing) was used for efficient search. Wang et al.~\cite{SELF:TPAMI:WLRLCC} proposed a weak label regularized sparse coding to enhance facial features and adopted the Locality-Sensitive Hash (LSH)~\cite{retrieval:LSH} to index the gallery.
  \item \emph{Product Quantization (PQ)}. Unlike the previous two approaches which require index vectors to be stored in main memory, PQ~\cite{retrieval:pq} is a compact discrete encoding method that can be used either for exhaustive search or inverted indexing search. In this work, we adopt product quantization for fast filtering.
\end{itemize}

Face search systems published in the literature have been mainly evaluated under closed-set protocols (Table~\ref{tab:fr}), which assume that the subject in the probe image is present in the gallery. However, in many large scale applications (e.g., surveillance and watch list scenarios), open-set search performance, where the probe subject may not be present in the gallery, is more relevant and appropriate.

A search operating in open-set protocol requires two steps: first determine if the identity associated with the face in the probe is present in the gallery, and if so find the top-$k$ most similar faces in the gallery. To address face search application requirements, several new protocols for unconstrained face recognition have been proposed, including the open-set identification protocol~\cite{faceretrieval:fusion} and the Benchmark of Large-scale Unconstrained Face Recognition (BLUFR) protocol~\cite{BLUFR}. However, even in these two protocols used on benchmark datasets, the gallery sizes are fairly small ($3,143$ and $1,000$ gallery images), due to the inherent small size of the LFW dataset. Table~\ref{tab:fr} shows that the largest face gallery size reported in the literature to date is about $1$M, which is not even close to being a representative of social media and forensic applications. To tackle these two limitations, we evaluate the proposed cascaded face search system with an $80$M face gallery under closed-set and open-set protocols.

\section{Face Search Framework}
Given a probe face image, a face search system aims to find the top-$k$ most similar face images in the gallery. To handle large galleries (e.g. tens of millions of images), we propose a cascaded face search structure, designed to speed up the search process while achieving acceptable accuracy ~\cite{Zhou19941539, fr:fastmatching}.

\begin{figure}[htpb]
  \centering
  \includegraphics[width=\linewidth]{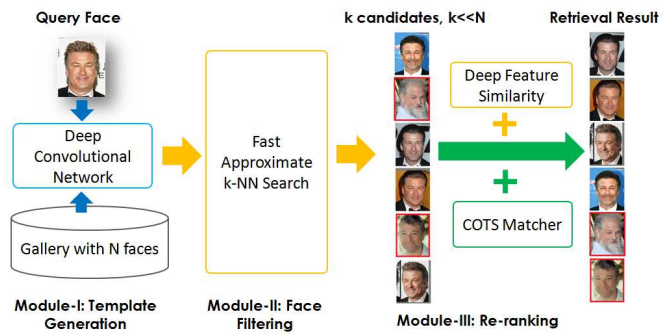}\\
  \caption{Illustration of the proposed large-scale face search system.}\label{fig:framework}
\end{figure}

Figure~\ref{fig:framework} outlines the proposed face search architecture consisting of three main steps: i) \emph{template generation} module which extracts features for the $N$ gallery faces offline as well as from the probe face; ii) \emph{face filtering} module which compares the probe representation against the gallery representations using product quantization to retrieve the top-$k$ most similar candidates ($k \ll N$); and (iii) \emph{re-ranking} module which fuses similarity scores of deep features with scores from a COTS face matcher to generate a new ordering of the $k$ candidates. These three modules are discussed in detail in the remainder of this section.

\subsection{Template Generation}\label{sec:template_generation}
Given a face image $I$, the template generator is a non-linear mapping function
\begin{equation}
\mathcal{F}(I)=\x \in \mathbb{R}^d
\end{equation}

\noindent which projects $I$ into a $d$-dimensional feature space. The discriminative ability of the template is critical for the accuracy of the search system. Given the impressive performance of deep learning techniques in various machine learning applications, particularly face recognition, we adopt deep learning for template generation.

\begin{figure}[htbp]
  \centering
    \includegraphics[width=\linewidth]{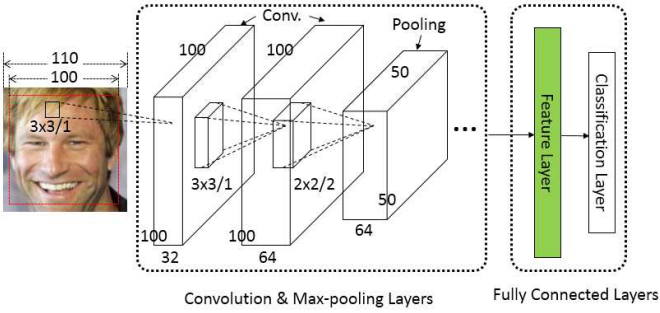}
  \caption{The proposed deep convolutional neural network (ConvNet).}\label{fig:architecture}
\end{figure}

The architecture of the proposed deep ConvNet (Fig.~\ref{fig:architecture}) is inspired by~\cite{self:wan2014, DB:CASIA}.
There are four main differences between the proposed network and the one in~\cite{DB:CASIA}: i) input to the network is color images instead of gray images; ii) a robust face alignment procedure; iii) an additional data argumentation step that randomly crops a $100\times100$ region from the $110\times110$ input color image; and iv) deleting the contrastive cost layer for computational efficiency (experimentally, this did not hinder recognition accuracy).

The proposed deep convolutional network has three major parts: i) convolution and pooling layers, ii) a feature representation layer, and iii) an output classification layer. For the convolution layers, we adopt a very deep architecture~\cite{dl:verydeep} ($10$ convolution layers in total) and filters with small supports ($3\times3$). The small filters reduce the total number of parameters to be learned, and the very deep architecture enhances the nonlinearity of the network. Based on the basic assumption that face images usually lie on a low dimensional manifold, the network outputs $320$ dimensional feature vector.

\begin{figure}[htbp]
  \centering
  \includegraphics[width=3.3in]{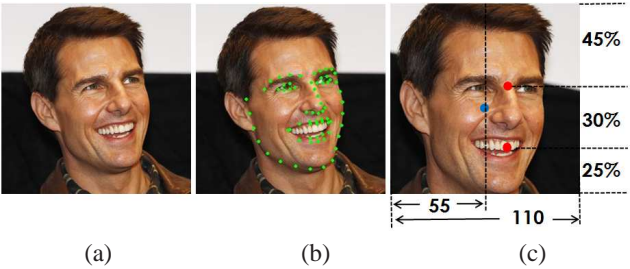} \\
  \makebox[1.1in]{(a)}
  \makebox[1.1in]{(b)}
  \makebox[1.1in]{(c)}
  \caption{A face image alignment example. The original image is shown in (a); (b) shows the 68 landmark points detected by the method in~\cite{kazemi2014one}, and (c) is the final aligned face image, where the blue circle was used to center the face image along the x-axis, and the red circles denote the two points used for face cropping.}\label{fig:alignment}
\end{figure}

The input layer accepts the RGB values of the aligned face image pixels. Faces are aligned as follows:
i) Use the  DLIB\footnote{\url{http://blog.dlib.net/2014/08/real-time-face-pose-estimation.html}} implementation of Kazemi and Sullivan's ensemble of regression trees method~\cite{kazemi2014one} to detect $68$ facial landmarks (see Fig.~\ref{fig:alignment});
ii) rotate the face in the image plane to make it upright based on the eye positions;
iii) find a central point on the face (the blue point in Fig.~\ref{fig:alignment}) by taking the mid-point between the leftmost and rightmost landmarks; the center points of the eyes and mouth (red points in Fig.~\ref{fig:alignment}) are found by averaging all the landmarks in the eye and mouth regions, respectively;
iv) center the faces in the x-axis, based on the central point (blue point);
v) fix the position along the y-axis by placing the eye center point at $45\%$ from the top of the image and the mouth center point at $25\%$ from the bottom of the image, respectively;
vi) resize the image to a resolution of $110 \times $110. Note that the computed midpoint is not consistent across pose. In faces exhibiting significant yaw, the computed midpoint will be different from the one computed in a frontal image, so facial landmarks are not aligned consistently across yaw.

We augment our training set using a couple of image transform operations: transformed versions of the input image are obtained by randomly applying horizontal reflection, and cropping random $100 \times 100$ sub-regions from the original $110 \times 110$ aligned faces images.

Following the input layer, there are $10$ convolutional layers, $4$ max-pooling layers, and $1$ average-pooling layer. To enhance the nonlinearity, every pair of convolutional layers is grouped together and connected sequentially. The first four groups of convolutional layers are followed by a max-pooling layer with a window size of $2\times2$ and a stride of $2$, while the last group of convolutional layers is followed by an average-pooling layer with window size $7\times7$. The dimensionality of the feature representation layer is the same as the number of filters in the last convolutional layer. As discussed in~\cite{DB:CASIA}, the ReLU~\cite{dl:alex2012} neuron produces a sparse vector, which is undesirable for a face representation layer. In our network, we use ReLU neurons~\cite{dl:alex2012} in all the convolutional layers, except the last one, which is combined with an average-pooling layer to generate a low dimensional face representation with a dimensionality of $320$.

Although multiple fully-connected layers are used in~\cite{dl:alex2012, self:wan2014}, in our network we directly feed the deep features generated by the feature layer to an $N$-way softmax (where $N=10,575$ is the number of subjects in our training set). We regularize the feature representation layer using dropout~\cite{dl:dropout}, keeping $60\%$ of the feature components as-is and randomly setting the remaining $40\%$ to zero during training.


We use a softmax loss function for our network, and train it using the standard back-propagation method. We implement the network using the open source cuda-convnet2\footnote{\url{https://code.google.com/p/cuda-convnet2/}} library. We set the weight decay of all layers to $5 \times 10^{-4}$. The learning rate for stochastic gradient descent (SGD) is initialized to $10^{-2}$, and gradually reduced to $10^{-5}$.

\subsection{Face Filtering}
Given a probe face $I$ and a template generation function $\mathcal{F}$, finding the top-$k$ most similar faces $\mathrm{C}_k(I)$ in the gallery $G$ is formulated as follows:
\begin{equation}
\mathrm{C}_k(I) = \mathrm{Rank}_k(\{ \mathcal{S}(\mathcal{F}(I), \mathcal{F}(J_i)) | J_{i=1, 2, \ldots, N} \in G \})
\end{equation}

\noindent where $N$ is the size of gallery $G$, $\mathcal{S}$ is a function, which measures the similarity of the probe face $I$ and the gallery image $J_i$, and $\mathrm{Rank}$ is a function that finds the top-$k$ largest values in an array. The computational complexity of naive face comparison functions is linear with respect to the gallery size $N$ and the feature dimensionality $d$. However, approximate nearest neighbor (ANN) algorithms, which improve runtime
without a significant loss in accuracy, have become popular for large galleries.

Various approaches have been proposed for ANN search. Hashing based algorithms use compact binary representations to conduct an exhaustive nearest neighbor search in Hamming space. Although multiple hash tables~\cite{retrieval:LSH} can significantly improve performance and reduce distortion, their performance degrades quickly with increasing gallery size in face recognition applications. Product quantization (PQ)~\cite{retrieval:pq}, where the feature template space is decomposed into a Cartesian product of low dimensional subspaces (each subspace is quantized separately) has been shown to achieve excellent search results~\cite{retrieval:pq}. Details of product quantization used in our implementation are described below.

Under the assumption that the dimensionality $d$ of the feature vectors is a multiple of $m$, where $m$ is an integer, any feature vector $\x \in \mathbb{R}^d$ can be written as a concatenation
$(\x^1, \x^2, \ldots, \x^m)$ of $m$ sub-vectors, each of dimension $d/m$. In the $i$-th subspace $\mathbb{R}^{d/m}$, given a sub-codebook $\mathcal{C}^i = \{ \c^i_{j=1,2,\ldots, z} | c^i_j \in \mathbb{R}^{d/m} \}$, where $z$ is the size of codebook, the sub-vector $\x^i$ can be mapped to a {codeword} $\c^i_j$ in the codebook $\mathcal{C}^i$,
with $j$ as the index value. The index $j$ can then be represented by a binary code with $\log_2(z)$ bits. In our system, each codebook is generated using the $k$-means clustering algorithm. Given all the $m$ sub-codebooks $\{ \mathcal{C}^1, \mathcal{C}^1, \ldots, \mathcal{C}^m \}$, the product quantizer of feature template $\x$ is
$$q(\x) = (q^1(\x^1), \ldots, q^m(\x^m))$$
where $q^j(\x^j) \in \mathcal{C}^j$ is the nearest sub-centroid of sub-vector $\x^j$ in $\mathcal{C}^j$, for $j = 1, 2, \dots, m$, and the quantizer $q(\x)$ requires $m\log_2(z)$ bits. Given another feature template $\y$, the asymmetric squared Euclidean distance between $\x$ and $\y$ is approximated by $$\mathcal{D}(\y, \x) = \| \y - q(\x) \|^2 = \sum_{j=1}^{m} \| \y^j - q^j(\x^j)\|^2$$
where $q^j(\x^j) \in \mathcal{C}^j$, and the distances $\| \y^j - q^j(\x^j)\|$ are pre-computed for each sub-vector of $\y^j, j=1,2, \ldots, m$ and each sub-centroid in $\mathcal{C}^j, j=1, 2, \ldots, m$. Since the distance computation requires $O(m)$ lookup and add operations~\cite{retrieval:pq}, approximate nearest neighbor search with product quantizers is fast, and significantly reduces the memory requirements with binary coding.

To further reduce the search time, a non-exhaustive search scheme was proposed in~\cite{retrieval:pq, retrieval:lopq} based on an inverted file system and a coarse quantizer; the query image is only compared against a portion of the image gallery, based on the coarse quantizer. Although a non-exhaustive search framework is essential for general image search problems based on local descriptors (where billions of local descriptors are indexed, and thousands of descriptors per query are typical), we found that non-exhaustive search significantly reduces face search performance when used with the proposed  feature vector.

Two important parameters in product quantization are the number of sub-vectors $m$ and the size of the sub-codebook $z$, which together determine the length of the quantization code: $m\log_2{z}$. Typically, $z$ is set to $256$. To find the optimal $m$, we empirically evaluate search accuracy and time per query for various values of $m$, based on a $1$ million face gallery and over $3,000$ queries.
We noticed that the performance gap between product quantization (PQ) and brute force search becomes small when the length of the quantization code is longer than $512$ bits ($m=64$). Considering search time, the PQ-based approximate search is an order of magnitude faster than the brute force search. As a trade-off between efficiency and effectiveness, we set the number of sub-vectors $m$ to $64$; The length of the quantization code is $64 \log_2(256) = 512$ bits.

Although we use product quantization to compute the similarity scores, we also need to pick a distance or similarity metric. We empirically evaluated cosine similarity, L$1$ distance, and L$2$ distance using a $5$M gallery. The cosine similarity achieves the best performance, although after applying L2 normalization, L2 distance has an identical performance.

\begin{figure*}[htbp]
\centering
  \includegraphics[width=6.4in]{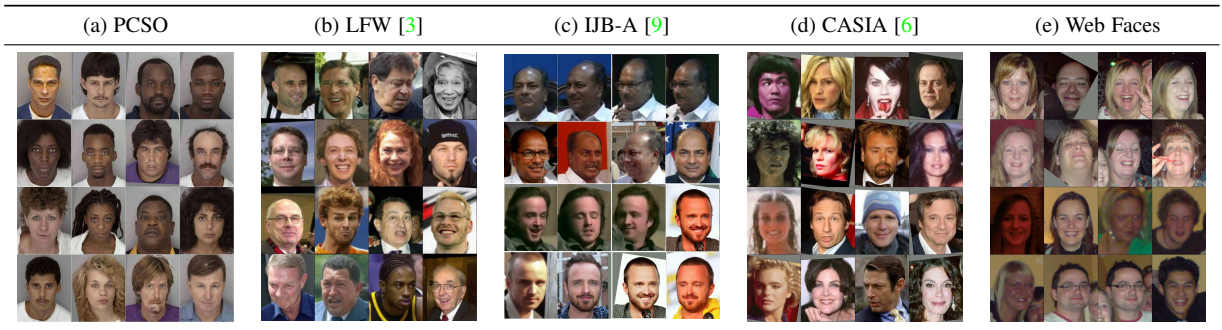}\\
  \caption{Examples of face images in five face datasets.}\label{fig:example_web}
\end{figure*}

\subsection{Re-Ranking}
After the short candidate list is acquired, the \emph{re-ranking module} aims to improve search accuracy by using several face matchers to re-rank the candidate list. In particular, given a probe face $I$ and the corresponding $k$ topmost nearest similar faces, $\mathrm{C}_k(I)$ returned from the filtering module, the $k$ candidate faces are re-ranked by fusing the similarity scores from $l$ different matchers. The ranking module is formulated as:
\begin{equation}
\mathrm{Sort}_{d}(\{ \mathrm{Fusion}(\mathcal{S}_{j=1,\ldots,l}(I, J_i)) | J_{i=1,\ldots, k} \in \mathrm{C}_k(I) \})
\end{equation}
\noindent where $\mathcal{S}_j$ is the $j$-th matcher, and $\mathrm{Sort}_{d}$ is a descending order sorting function. In general, there is a trade-off between accuracy and computational cost when using multiple face recognition approaches. To make our system simple yet effective, we set $l=2$ and generate the final similarity score using the sum-rule fusion~\cite{fusion:occ} of the cosine similarity from the proposed deep network, and the scores generated by a stat-of-the-art COTS face matcher with z-score normalization~\cite{zscore}.

The main benefits of combining the proposed deep features and a COTS matcher are threefold:
1) the cosine similarities can be easily acquired from the fast filtering module;
2) an important guideline in fusion is that the matchers should have some diversity~\cite{fusion:occ, fusion:rule}. We noticed that the set of impostor face images that are incorrectly assigned high similarity scores by deep features and COTS matcher do not overlap.
3) COTS matchers are widely deployed in many real world applications~\cite{fvrt:2014}, so the proposed cascade fusion scheme can be easily integrated in existing applications to improve scalability and performance.

\subsubsection{Impact of Size of Candidate Set ($k$)}
In the proposed cascaded face search system, the size of candidate list $k$ is a key parameter. In general, we expect the optimal value of $k$ to be related to the gallery size $N$ (a larger gallery would require a larger candidate list to maintain good search performance). We evaluate the relationship between $k$ and $N$ by computing the mean average precision (mAP) as the gallery size ($N$) from $100$K to $5$M and the size of candidate list ($k$) from $50$ to $100$K.
\begin{figure}[htbp]
\centering
  \includegraphics[width=3in]{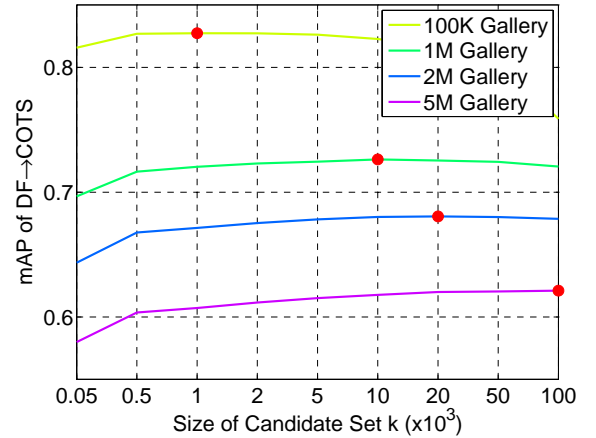}
  \caption{Impact of candidate set size $k$ as a function of the size of the gallery ($N$) on the search performance. Red points mark the optimal value of the candidate set ($k$) size for each gallery size.}\label{fig:exp_k}
\end{figure}

Fig.~\ref{fig:exp_k} shows the search performance, as expected, decreases with increasing gallery size. Further, if we increase $k$ for a fixed $N$, search performance will initially increase, then drop off when $k$ gets too large.  We find that the optimal candidate set size $k$ scales linearly with the size of the gallery $N$. Because the plots in Fig~\ref{fig:exp_k} flatten out, a near optimal value of $k$ (e.g., $k=0.01$N) can drastically reduce the candidate list with only a very small loss in accuracy.

\subsubsection{Fusion Method}
Another important issue for the proposed cascaded search system is the fusion of similarity scores from deep features (DF) and COTS. We empirically evaluated the following strategies:
\begin{itemize}
  \item {\bf DF$+$COTS:} Score-level fusion of similarities based on deep features and the COTS matcher, without any filtering.
  \item	{\bf DF$\rightarrow$COTS:} Filter the gallery using deep features, then re-rank the candidate list based on score-level fusion between the deep features and the COTS scores.
  \item {\bf DF$\rightarrow$COTS$_\mathrm{only}$:} Only use the similarity scores of COTS matcher to rank the $k$ candidate faces.
  \item {\bf DF$\rightarrow$COTS$_\mathrm{rank}$:} Rank all the $k$ candidate faces with COTS and deep features scores separately, then combine the two ranked lists using rank-level fusion. This is useful when the COTS matcher does not report similarity scores.
\end{itemize}
\begin{figure}
\centering
  \includegraphics[width=3in]{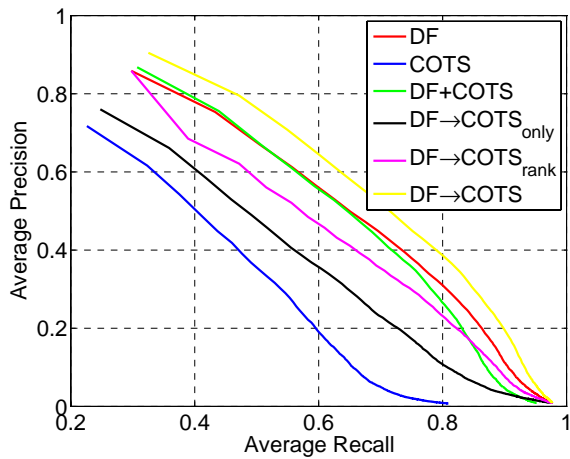}\\
  \caption{Comparison of fusion strategies based on a $1$M gallery. }\label{fig:fusion}
\end{figure}
We evaluated the different fusion methods on a $1$M face gallery. The \emph{average precision} vs. \emph{average recall} curves of these four fusion strategies are shown in Fig.~\ref{fig:fusion}. As a base line, we also show the performance of just using DF and COTS alone. The fusion scheme (DF$\rightarrow$COTS) consistently outperforms the other fusion methods as well as simply using DF and COTS alone. Note that omitting the filtering step results does not perform as well as the cascaded approach, which is consistent with results in the previous section: when $k$ is too large (e.g. $k=N$), the search accuracy decreases.

\begin{table*}[htbp]
\centering
\caption{Performance of various face recognition methods on the standard LFW verification protocol.}\label{tab:lfw_standard}
\begin{tabular}{l|rlll}
\toprule
Method                      & \#Nets & Training Set (private or Public face dataset) & Training Setting & Mean accuracy $\pm$ sd \\ \midrule
DeepFace~\cite{dl:deepface} & 1 & $4.4$ million images of $4,030$ subjects, Private & cosine & 95.92\%$\pm$0.29\%\\
DeepFace  & 7 & $4.4$ million images of $4,030$ subjects, Private    & unrestricted, SVM    & 97.35\%$\pm$0.25\% \\
DeepID2~\cite{dl:deepID2} & 1 & $202,595$ images of $10,117$ subjects, Private & unrestricted, Joint-Bayes & 95.43\% \\
DeepID2 & 25    & $202,595$ images of $10,117$ subjects, Private    & unrestricted, Joint-Bayes & $99.15\pm0.15\%$ \\
DeepID3~\cite{dl:deepID3} & 50 & $202,595$ images of $10,117$ subjects, Private & unrestricted, Joint-Bayes & $99.53\pm0.10\%$ \\
Face$++$~\cite{dl:face++} & 4  & $5$ million images of $20,000$ subjects, Private & L2 & $99.50\pm0.36\%$\\
FaceNet~\cite{dl:facenet} & 1  & $100 \sim 200$ million images of $8$ million subjects, Private & L2 & $99.63\pm0.09\%$ \\
Tencent-BestImage~\cite{dl:tencent}  & 20 & $1,000,000$ images of $20,000$ subjects, Private & Joint-Bayes & $99.65\pm0.25\%$ \\
Li \etal~\cite{DB:CASIA} & 1  & $494,414$ images of $10,575$ subjects, Public & cosine & 96.13\%$\pm$0.30\%\\
Li \etal   & 1  & $494,414$ images of $10,575$ subjects, Public &unrestricted, Joint-Bayes & 97.73\%$\pm$0.31\% \\
Human, funneled                 & N/A & N/A   & N/A & 99.20\%\\
COTS                    & N/A   & N/A   & N/A & 90.35\%$\pm$1.30\%\\
\hline
Proposed Deep Model & 1  & $494,414$ images of $10,575$ subjects, Public & cosine & 96.95\%$\pm$1.02\% \\
Proposed Deep Model & 7  & $494,414$ images of $10,575$ subjects, Public & cosine & 97.52\%$\pm$0.76\% \\
Proposed Deep Model & 1  & $494,414$ images of $10,575$ subjects, Public & unrestricted, Joint-Bayes & 97.45\%$\pm$0.99\%  \\
Proposed Deep Model & 7  & $494,414$ images of $10,575$ subjects, Public & unrestricted, Joint-Bayes & 98.23\%$\pm$0.68\% \\
\bottomrule
\end{tabular}
\end{table*}

\section{Face Datasets}\label{sec:database}
We use four web face datasets and one mugshot dataset in our experiments: PCSO, LFW~\cite{DB:LFWTech}, IJB-A~\cite{db:janus}, CASIA-WebFace~\cite{DB:CASIA} (abbreviated as ``CASIA" in the following sections), and general web face images, referred to as ``Web Faces", which we downloaded from the web to augment the gallery. We briefly introduce these datasets, and show example face images from each dataset (Fig.~\ref{fig:example_web}).
\begin{itemize}
  \item \textbf{PCSO:} This dataset is a subset of a larger collection of mugshot images acquired from the Pinellas County Sheriffs Office (PCSO) dataset, which contains $1,447,607$ images of $403,619$ subjects.
  \item \textbf{LFW}~\cite{DB:LFWTech}: The LFW dataset is a collection of $13,233$ face images of $5,749$ individuals, downloaded from the web. Face images in this dataset contain significant variations in pose, illumination, and expression. However, the face images in this dataset were selected on the bias that they could be detected by the Viola-Jone detector~\cite{DB:LFWTech,Viola}.
  \item \textbf{IJB-A}~\cite{db:janus} IARPA Janus Benchmark-A (IJB-A) contains $500$ subjects with a total of $25,813$ images ($5,399$ still images and $20,414$ video frames). Compared to the LFW dataset, the IJB-A dataset is more challenging due to: i) full pose variation making it challenging to detect all the faces using a commodity face detector, ii) a mix of images and videos, and iii) wider geographical variation of subjects. To make evaluation of face recognition methods feasible in the absence of automatic face detection and landmarking methods  for images with full-pose variations, ground-truth eye, nose and face locations are provided with the IJB-A dataset (and used in our experiments when needed). Fig.~\ref{fig:example_web} (c) shows the images of two different subjects in the IJB-A dataset, captured in various conditions (video/photo, indoor/outdoor, pose, expression, illumination).
  \item \textbf{CASIA}~\cite{DB:CASIA} dataset provides a large collection of labeled (based on subject names) training set for deep learning networks. It contains $494,414$ images of $10,575$ subjects.
  \item \textbf{Web Faces} To evaluate the face search system on a large-scale gallery, we used a crawler to automatically download millions of web images, which were filtered to only include images with faces detectable by the OpenCV implementation of the Viola-Jones face detector~\cite{Viola}. A total of $80$ million face images were collected in this manner, which were used to augment the gallery in our experiments.
\end{itemize}

\section{Face Recognition Evaluation}
In this section, we first evaluate the proposed deep models on a mugshot dataset (PCSO), then we evaluate the performance of the proposed deep model on two publicly available unconstrained face recognition benchmarks (LFW~\cite{DB:LFWTech} and IJB-A~\cite{db:janus}) to establish its performance relative to the state of the art.

\subsection{Mugshot Evaluation}
We evaluate the proposed deep model using the PCSO mugshot dataset. Some example mugshots are shown in Fig.~\ref{fig:example_web} (a). Images are captured in constrained environments with a frontal view of the face. We compare the performance of our deep features with a COTS face matcher. The COTS matcher is designed to work with mugshot-style images, and is one of the top performers in the 2014 NIST FRVT~\cite{fvrt:2014}.

Since mugshot dataset is qualitatively different from the CASIA~\cite{db:janus} dataset that we used to train our deep network, similar to~\cite{icb2015}, we first retrained the network with a mugshot training set taken from the full PCSO dataset, consisting of $471,130$ images of $29,674$ subjects. Then, we compared the performance of deep features with the COTS matcher on a test subset of the PCSO dataset containing $89,905$ images of $13,665$ subjects, which contains no overlapping subjects with the training set. We evaluate performance in the verification scenario, and make a total of about $340$K genuine pairwise comparisons and over $4$ billion impostor pairwise comparisons. The experimental results are shown in Table~\ref{tab:pcso}. We observe that the COTS matcher outperforms the deep features consistently, especially at low false accept rates (FAR) (e.g. 0.01\%). However, a simple score-level fusion between the deep features and COTS scores results in improved performance.

\begin{table}[htbp]
\scriptsize
\centering
\caption{Performance of face verification on a subset of the PCSO dataset ($89,905$ images of $13,666$ subjects). There are about $340$K genuine pairs and over $4$ billion imposter pairs.}\label{tab:pcso}
\begin{tabular}{l|ccc}
\toprule
              & TAR@FAR=0.01\% & TAR@FAR=0.1\% & TAR@FAR=1\%\\ \midrule
COTS          & 0.985    & 0.993     & 0.997 \\
Deep Features & 0.935    & 0.977     & 0.993 \\ \cmidrule{1-1}
DF + COTS     & 0.992    & 0.996     & 0.997 \\
\bottomrule
\end{tabular}
\end{table}

\subsection{LFW Evaluation}
While mugshot data is of interest in some applications, many others require handling more difficult, unconstrained face images. In this section, we evaluate the proposed deep models on a more difficult dataset, the  LFW~\cite{DB:LFWTech} unconstrained face dataset, using two protocols: the standard LFW~\cite{DB:LFWTech} protocol and the BLUFR protocol~\cite{BLUFR}.

\subsubsection{Standard Protocol}\label{sec:lfw_protocol}
The standard LFW evaluation protocol defines $3,000$ pairs of genuine comparisons and $3,000$ pairs of impostor comparisons, involving $7,701$ images of $4,281$ subjects. These $6,000$ face pairs are divided into $10$ disjoint subsets for cross validation, with each subset containing $300$ genuine pairs and $300$ impostor pairs. We compare the proposed deep model with several state-of-the-art deep models: DeepFace~\cite{dl:deepface}, DeepID2~\cite{dl:deepID2}, DeepID3~\cite{dl:deepID3}, Face++~\cite{dl:face++}, DeepNet~\cite{dl:facenet}, Tencent-BestImage~\cite{dl:tencent}, and Li~\etal~\cite{DB:CASIA}. Additionally, we report the performance of a state-of-the-art commercial face matcher (COTS), as well as human performance on ``funneled" LFW images~\cite{align:funnel}.

Based on the experimental results shown in Table~\ref{tab:lfw_standard}, we can make the following observations: (i) the COTS matcher performs poorly relative to the deep learning based algorithms. This is to be expected since most COTS matchers have been trained to handle face images captured in constrained environments, e.g. mugshot or driver license photos. (ii) The superior performance of deep learning based algorithms can be attributed to (a) large number of training images ($> 100$K), (b) data augmentation methods, e.g., use of multiple deep models, and (c) supervised learning algorithms, such as Joint-Bayes~\cite{ml:jointbayes}, used to learn a verification model for a pair of faces in the training set.

To generate multiple deep models, we cropped $6$ different sub-regions from the aligned face images (by centering the positions of the left-eye, right-eye, nose, mouth, left-brow, and right-brow) and trained six additional networks. As a result, by combining seven models together and using Joint-Bayes~\cite{ml:jointbayes}, the performance of our deep model can be improved to $98.23\%$ from $96.95\%$ for a single network using the cosine similarity. Despite using only publicly available training data, the performance of our deep model is highly competitive with state-of-the-art on the standard LFW protocol (see Table~\ref{tab:lfw_standard}).

\subsubsection{BLUFR Protocol}
It has been argued in the literature that the standard LFW evaluation protocol is not appropriate for real-world face recognition systems, which require high true accept rates (TAR) at low false accept rates ( FAR $=0.1\%$). A number of new protocols for unconstrained face recognition have been proposed, including the open-set identification protocol~\cite{faceretrieval:fusion} and the Benchmark of Large-scale Unconstrained Face Recognition (BLUFR) protocol~\cite{BLUFR}. In this experiment, we follow the BLUFR protocol, which defines $10$-fold cross-validation \emph{face verification} and \emph{open-set identification} tests, with corresponding training sets for each fold.

For \emph{face verification}, in each trial, the test set contains the $9,708$ face images of $4,249$ subjects, on average. As a result, over $47$ million face comparison scores need to be computed in each trial.  For \emph{open-set identification}, the dataset in the previous verification task ($9,708$ face images of $4,294$ subjects) is randomly partitioned into three subsets: gallery set, genuine probe set, and impostor probe set. In each trial, $1,000$ subjects from the test set are randomly selected to constitute the gallery set; a single image per subject is put in the gallery. After the gallery is selected, the remaining images from the $1,000$ selected subjects are used to form the genuine probe set, and all other images in the test set are used as the impostor probe set. As a result, in each trial, the genuine probe set contains $4,350$ face images of $1,000$ subjects, the impostor probe set contains $4,357$ images of $3,249$ subjects, on average, and the gallery set contains $1,000$ images.

Following the protocol in~\cite{BLUFR}, we report the verification rate (VR) at FAR $=0.1\%$ for the \emph{face verification} and the detection and identification rate (DIR) at Rank-$1$ corresponding to an FAR of $1\%$ for {open-set identification}. As yet, only a few other deep learning based algorithms have reported their performance using this protocol. We report the published results on this protocol, along with the performance of our deep network, and a state of the art COTS matcher in Table~\ref{tab:lfw_blufr}.

\begin{table}[htbp]
\scriptsize
\centering
\caption{
Performance of various face recognition methods using the BLUFR LFW protocol reported as Verification Rate (VR) and Detection and Identification Rate (DIR).}\label{tab:lfw_blufr}
\begin{tabular}{l|l|c|c}
\toprule
\multirow{1}{*}{Method}          & \multirow{1}{*}{Training Setting}  & VR          & DIR@FAR=1\%   \\
                                 &                                    & @FAR=0.1\%  & Rank=1        \\
\midrule
HDLBP+JB~\cite{BLUFR}            & Joint-Bayes                        & 41.66\% & 18.07\% \\
HDLBP+LDA~\cite{BLUFR}           & LDA                                & 36.12\% & 14.94\% \\
Li~\etal~\cite{DB:CASIA}         & Joint-Bayes                        & 80.26\% & 28.90\% \\
COTS                             & N/A                                & 58.56\% & 36.44\% \\ \hline
Proposed Deep Model              & \#Nets $=1$, Cosine                    & 83.39\% & 46.31\% \\
Proposed Deep Model              & \#Nets $=7$, Cosine                    & 87.65\% & 56.27\% \\
\bottomrule
\end{tabular}
\end{table}

We notice that the verification rates at low FAR ($0.1\%$) under the BLUFR protocol are much lower than the accuracies reported on the standard LFW protocol. For example, the performance of the COTS matcher is only $58.56\%$ under the BLUFR protocol compared to $90.35\%$ in the standard LFW protocol. This indicates that the performance metrics for the BLUFR protocol are much harder as well as realistic than those of the standard LFW protocol. The deep learning based algorithms still perform better than the COTS matcher, as well as the high-dimensional LBP based features. Using cosine similarity and a single deep model, our method achieves better performance ($83.39\%$) than the one in~\cite{DB:CASIA}, which indicates that our modifications to the network design (using RGB input, random cropping, and improved face alignment) does improve the recognition performance. Our performance is further improved to $87.65\%$ when we fuse $7$ deep models. In this experiment, Joint-Bayes approach~\cite{ml:jointbayes} did not improve the performance. In the open-set recognition results, our single deep model achieves a significantly better performance ($46.31\%$) than the previous best reported result of $28.90\%$~\cite{DB:CASIA} and the COTS matcher ($36.44\%$).

\subsection{IJB-A Evaluation}
The IJB-A dataset~\cite{db:janus} was released in an attempt to push the frontiers of unconstrained face recognition. Given that the recognition performance on the LFW dataset was getting saturated and the deficiencies in the LFW protocols, the IJB-A dataset contains more challenging face images and defines both verification and identification (open and close sets) protocols. The basic protocol consists of 10-fold cross-validation on pre-defined splits of the dataset, with a disjoint training set defined for each split.
\begin{figure}[htbp]
    \centering
    \includegraphics[width=3.4in]{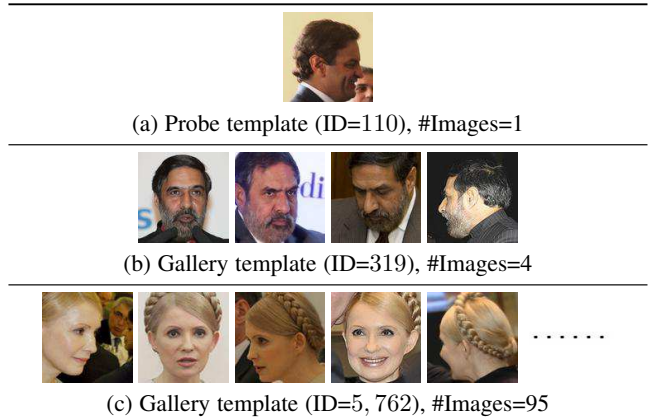}
    \caption{Examples of probe/gallery ``templates" in the first folder of IJB-A protocol in 1:N face search.}\label{fig:tempaltes}
\end{figure}

One unique aspect of the IJB-A evaluation protocol is that it defines ``templates," consisting of one or more images (still images or video frames), and defines set-to-set comparisons, rather than using face-to-face comparisons. Fig.~\ref{fig:tempaltes} shows examples of templates in the IJB-A protocol (one per row), with varying number of images per template. In particular, in the IJB-A evaluation protocol the number of images per template ranges from a single image to a maximum of $202$ images.  Both the search task (1:N search) and verification task (1:1 matching) are defined in terms of comparisons between templates (consisting of several face images), rather than single face images.

The verification protocol in IJB-A consists of $10$ sets of pre-defined comparisons between templates (groups of images). Each set contains about $11,748$ pairs of templates ($1,756$ genuine plus $9,992$ impostor pairs), on average.
For the search protocol, which evaluates both closed-set and open-set search performance, $10$ corresponding gallery and probe sets are defined, with both the gallery and probe sets consisting of templates. In each search fold, there are about $1,187$ genuine probe templates, $576$ impostor probe templates, and $112$ gallery templates, on average.

\begin{figure}
\centering
  \includegraphics[width=3.4in]{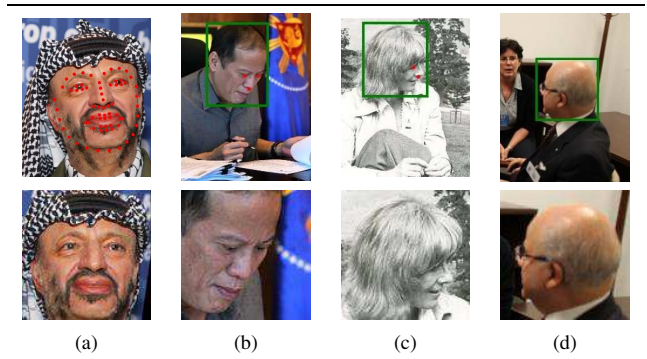}
  \caption{Examples of web images in the IJB-A dataset with overlayed landmarks (top row), and the corresponding aligned face images (bottom row); (a) example of a well-aligned image obtained using automatically detected landmarks by DLIB~\cite{kazemi2014one}; (b), (c), and (d) examples of poorly-aligned images with 3, 2, and 0 ground-truth landmarks provided in IJB-A, respectively. DLIB fails to output landmarks for (b)-(d). The web images in the top row have been cropped around the relevant face regions from the original images.} \label{fig:ijba-alignment}
\end{figure}

Given an image or video frame from the IJB-A dataset, we first attempt to automatically detect $68$ facial landmarks with DLIB. If the landmarks are successfully detected, we align the detected face using the alignment method proposed in Section~\ref{sec:template_generation}. We call the images with automatically detected landmarks \emph{well-aligned image}s. If the landmarks cannot be automatically detected, as is the case for profile faces or when only the back of the head is showing (Fig.~\ref{fig:ijba-alignment}), we align the face based on the ground-truth landmarks provided with the IJB-A protocol. All possible ground truth landmarks (left eye, right eye, and nose tip) may be visible in every image, and so the M-Turk workers who manually marked the landmarks skipped the missing ones. For example, in faces exhibiting a high degree of yaw, only one eye is typically visible. If all the three landmarks are available, we estimate the mouth position and align the face images using the alignment method in Section~\ref{sec:template_generation}; otherwise, we directly crop a square face region using the provided ground-truth face region. We call images for which the automatic landmark detection fails \emph{poorly-aligned image}s. Fig.~\ref{fig:ijba-alignment} shows some examples from these two categories in the IJB-A dataset.

\newcommand{\mtablew}{0.70in}
\newcommand{\mimgw}{0.65in}

The IJB-A protocol allows participants to perform training for each fold. Since the IJB-A dataset is qualitatively different from the CASIA dataset that we used to train our network, we retrain our deep model using the IJB-A training set. The final face representations consists of a concatenation of the deep features from five deep model trained just on the CASIA dataset and one re-trained (on the IJB-A training set following the protocol) deep model. We then reduce the dimensionality of the combined face representation to $100$ using PCA.

Since all the IJB-A comparisons are defined between sets of faces, we need to determine an appropriate set-to-set comparison method. We choose to prioritize \emph{well-aligned image}s, since they are most consistent with the data used to train our deep models. Our set-to-set comparison strategy is to check if there are one or more \emph{well-aligned image}s in a template. If so, we only use the \emph{well-aligned image}s for the set comparison, we call the corresponding template \emph{well-aligned template}s. Otherwise we use the \emph{poorly-aligned image}s, with naming the corresponding template \emph{poorly-aligned template}s. The pairwise face-to-face similarity scores are computed using the cosine similarity, and the average score over the selected subset of images is the final set-to-set similarity score.

\begin{figure*}[htbp]
    \centering
    \includegraphics[width=6.4in]{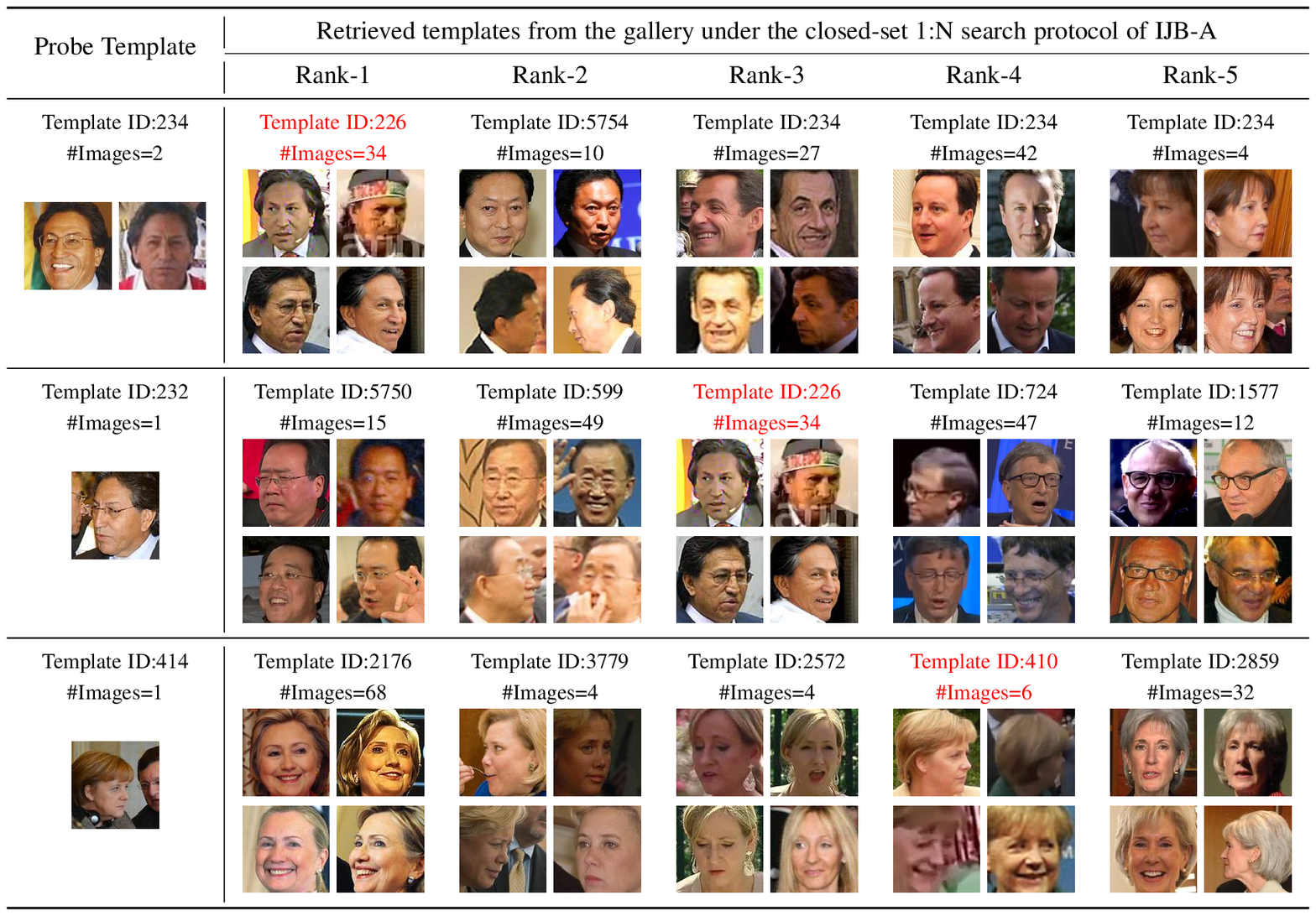}
    \vspace{-0.1in}
    \caption{Examples of face search in first fold of the IJB-A closed-set 1:N search protocol, using ``templates." The first column contains the probe templates, and the following $5$ columns contain the corresponding top-$5$ ranked gallery templates, where \textcolor{red}{red} text highlights the correct mated gallery template. There are $112$ gallery templates in total; only a subset (four) of the gallery images for each template are shown.}\label{fig:ijba-search}
\end{figure*}

\begin{table*}[htbp]
    \caption{Recognition accuracies under the IJB-A protocol. Results for GOTS and OpenBR are taken from~\cite{db:janus}. Results reported are the average $\pm$ standard deviation over the $10$ folds specified in the IJB-A protocol.}
    \label{tab:ijba_perf}
    \begin{tabular}{lccccccc}
        \toprule
            & \multicolumn{3}{c}{TAR @ FAR (verification)} & \multicolumn{2}{c}{CMC (closed-set search)} & \multicolumn{2}{c}{FNIR @ FPIR (open-set search):} \\
            \cmidrule(lr){2-4} \cmidrule(lr){5-6} \cmidrule(lr){7-8}
        Algorithm    & 0.1 & 0.01 & 0.001 & Rank-1 & Rank-5 & 0.1 & 0.01 \\
        \midrule
        GOTS & $0.627 \pm 0.012$ & $ 0.406 \pm 0.014$ & $0.198 \pm 0.008$ &  $ 0.443 \pm 0.021$  & $0.595 \pm 0.020$
                & $0.765 \pm 0.033$ & $0.953 \pm 0.024$ \\
        OpenBR & $0.433 \pm 0.006$ & $0.236 \pm 0.009$ & $0.104 \pm 0.014$ & $0.246 \pm 0.011$ & $0.375 \pm 0.008$ & $0.851 \pm0.028$ & $0.934 \pm 0.017$ \\
        Proposed Deep Model & $0.895 \pm 0.013$ & $0.733 \pm 0.034$ & $0.514 \pm 0.060$ & $0.820 \pm 0.024$ & $0.929 \pm 0.013$ & $0.387 \pm0.032$ & $0.617 \pm 0.063$ \\
        \bottomrule
    \end{tabular}
\end{table*}

\begin{figure}
\centering
  \includegraphics[width=3in]{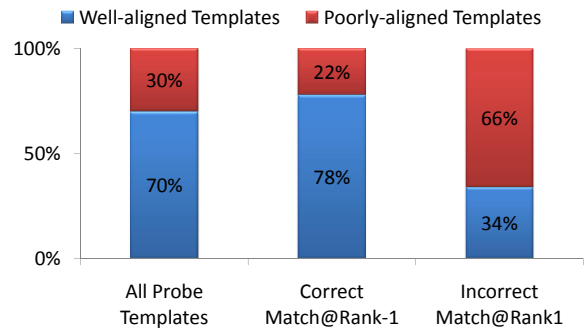}\\
  \caption{Distribution of well-aligned templates and poorly-aligned templates in 1:N search protocol of IJB-A, averaged over $10$ folds. Correct Match@Rank-$1$ means that the mated gallery template is correctly retrieved at rank $1$. Well-aligned images use the landmarks automatically detected by DLIB~\cite{kazemi2014one}. Poorly-aligned images mainly consist of side-views of faces. We align these images using the three ground-truth landmarks where available, or else by cropping the entire face region. }\label{fig:stats}
\end{figure}

In terms of evaluation, verification performance is summarized using True Accept Rates (TAR) at a fixed False Accept Rate (FAR). The TAR is defined as the fraction of genuine templates correctly accepted at a particular threshold, and FAR is defined as the fraction of impostor templates incorrectly accepted at the same threshold. Closed-set recognition performance is evaluated based on the Cumulative Match Characteristic (CMC) curve, which computes the fraction of genuine samples retrieved at or below a specific rank. Open-set recognition performance is evaluated using the False Positive Identification Rate (FPIR), and the False Negative Identification Rate (FNIR), where FPIR is the fraction of impostor probe images accepted at a given threshold, while FNIR is the fraction of genuine probe images rejected at the same threshold. Key results of the proposed method, along with the baseline results reported in~\cite{db:janus} are shown in Table~\ref{tab:ijba_perf}. Our deep network based method performs significant better than the two baselines at all evaluated operating points. Fig.~\ref{fig:ijba-search} shows three sets of face search results. We failed to find the mated templates at rank $1$ for the third probe template. A template containing a single poorly-aligned image is much harder to recognize than the templates containing one or more well-aligned images. Fig.~\ref{fig:stats} shows the distribution of well-aligned images and poorly-aligned images in probe templates. Compared to the distribution of poorly aligned templates in the overall dataset, we fail to recognize a disproportionate number of templates containing only poorly-aligned face images at rank $1$.

\section{Large-scale Face Search}
In this section, we evaluate our face search system using an $80$M gallery. The test datasets we use include LFW and IJB-A data, but now we do not follow the protocols associated with these two datasets, and instead use those images as the mated portion of a retrieval database with an enhanced gallery. We report search results, both under open-set and closed-set protocols, with increasing size of the gallery up to 80M faces. We evaluate the following three face search schemes:
\begin{itemize}
  \item {\bf Deep Features (DF):} Use our deep features and product quantization (PQ) to directly retrieve the top-$k$ most similar faces to the probe (no re-ranking step).
  \item {\bf COTS:} Use a state-of-the-art COTS face matcher to compare the probe image with each gallery face, and output the top-$k$ most similar faces to the probe (no filtering step).
  \item {\bf DF$\rightarrow$COTS:} Filter the gallery using deep features and then re-rank the $k$ candidate faces by fusing cosine similarities computed from deep features with the COTS matcher's similarity scores.
\end{itemize}

For closed-set face search, we assume that the probe always has at least one corresponding face image in the gallery. For open-set face search, given a probe we first decide whether a corresponding image is present in the gallery. If it is determined that the probe's identity is represented in the gallery, then we return the search results for the probe image. For open-set performance evaluation, the probe set consists of two groups: i) genuine probe set that has mated images in the gallery set, and ii) impostor probe set that has no mated images in the gallery set.

\subsection{Search Dataset}
We construct a large-scale search dataset using the four web face datasets introduced in Section~\ref{sec:database}. The dataset consists of five parts, as shown in Table~\ref{tab:web_ret}: 1) \emph{training set}, which is used to train our deep network; 2) \emph{genuine probe set}, the probe set which has corresponding gallery images; 3) \emph{mate set}, the part of the gallery containing the same subjects as the \emph{genuine probe set}; 4) \emph{impostor probe set}, which has no overlapping subjects with the \emph{genuine probe set}; 5) \emph{background set}, which has no identity labels and is simply used as background images to enlarge the gallery size.

We use the LFW and IJB-A datasets to construct the \emph{genuine probe set} and corresponding \emph{mate set}.
For the LFW dataset, we first remove all the subjects who have only a single image, resulting in $1,507$ subjects with 2 or more images. For each of these subjects, we take half of the images for the \emph{genuine probe set} and use the remaining images for the \emph{mate set} in the gallery. We repeat this process $10$ times to generate $10$ groups of probe and mate sets. To construct the \emph{impostor probe set}, we collect $4,000$ images from the LFW subjects with only a single image. For the IJB-A dataset, a similar process is adopted to generate $10$ groups of probe and mate sets. To build a large-scale \emph{background set}, we use a crawler to download millions of web images from the Internet, then filter them to only include those with faces detectable by the OpenCV implementation of the Viola-Jones face detector. By combining \emph{mate set} and \emph{background set}, we compose an $80$ million web face gallery. More details are shown in Table~\ref{tab:web_ret}.
\begin{table}[htbp]
\centering
\caption{Large-scale web face search dataset overview.}\label{tab:web_ret}
    \begin{tabular}{llrr}
    \toprule
                       & Source                   & \# Subjects  & \# Images  \\ \midrule
    Training Set       & CASIA~\cite{DB:CASIA}    & 10,575       & 494,414           \\ \midrule
    \multicolumn{4}{c}{LFW based probe and mate sets} \\
    Genuine Probe Set  & LFW~\cite{DB:LFWTech}    & 1,507        & 3,370             \\
    Mate Set           & LFW~\cite{DB:LFWTech}    & 1,507        & 3,845             \\ \midrule
    \multicolumn{4}{c}{IJB-A based probe and mate sets} \\
    Genuine Probe Set  & IJB-A~\cite{DB:LFWTech}  & 500          & 10,868            \\
    Mate Set           & IJB-A~\cite{DB:LFWTech}  & 500          & 10,626            \\ \midrule
    Impostor Probe Set & LFW~\cite{DB:LFWTech}    & 4,000        & 4,000             \\
    Background Set     & Web Faces                & N/A           & 80,000,000        \\
    \bottomrule
    \end{tabular}
\end{table}

\subsection{Performance Measures}\label{sec:performance}
We evaluate face search performance in terms of \emph{precision}, the fraction of the search set consisting of mated face images and \emph{recall}, the fraction of all mated face images for a given probe face that were returned in the search results. Various trade-offs between precision and recall are possible (for example, high recall can be achieved by returning a large search set, but a large search set will also lead to lower precision), so we summarize overall closed-set face search performance using \emph{mean Average Precision} (mAP). The mAP measure is widely used in image search applications, and is defined as follows: given a set of $n$ probe face images $Q=\{ \x_{q}^1, \x_{q}^2, \ldots, \x_{q}^n \}$ and a gallery set with $N$ images, the \emph{average precision} of $\x_{q}^i$ is:
\begin{equation}\label{eq:avgPres}
    \mathrm{avgP}(\x_q^i) = \sum_{k=1}^{N} P(k) \times [R(k) - R(k-1)]
\end{equation}

\noindent where $P(k)$ is \emph{precision} at the position $k$ and $R(k)$ is \emph{recall} at the position $k$ with $R(0)=0$. The mean Average Precision (mAP) of the entire probe set is then:
$$\mathrm{mAP}(Q) = \mathrm{mean}(\mathrm{avgP}(\x_q^i)), i=1,2,\ldots, n$$

\noindent In the open-set scenario, we evaluate search performance as a trade-off between mean average precision (mAP) and false accept rate (FAR) (the fraction of impostor probe images which are not rejected at a given threshold). Given a genuine probe, its \emph{average precision} is set to $0$ if it is rejected at a given threshold, otherwise, its \emph{average precision} is computed with Eq.~\ref{eq:avgPres}. A basic assumption in our search performance evaluation is that none of the query images are present in the $80$M downloaded web faces.

\begin{figure}[htbp]
  \centering
  \includegraphics[width=3.4in]{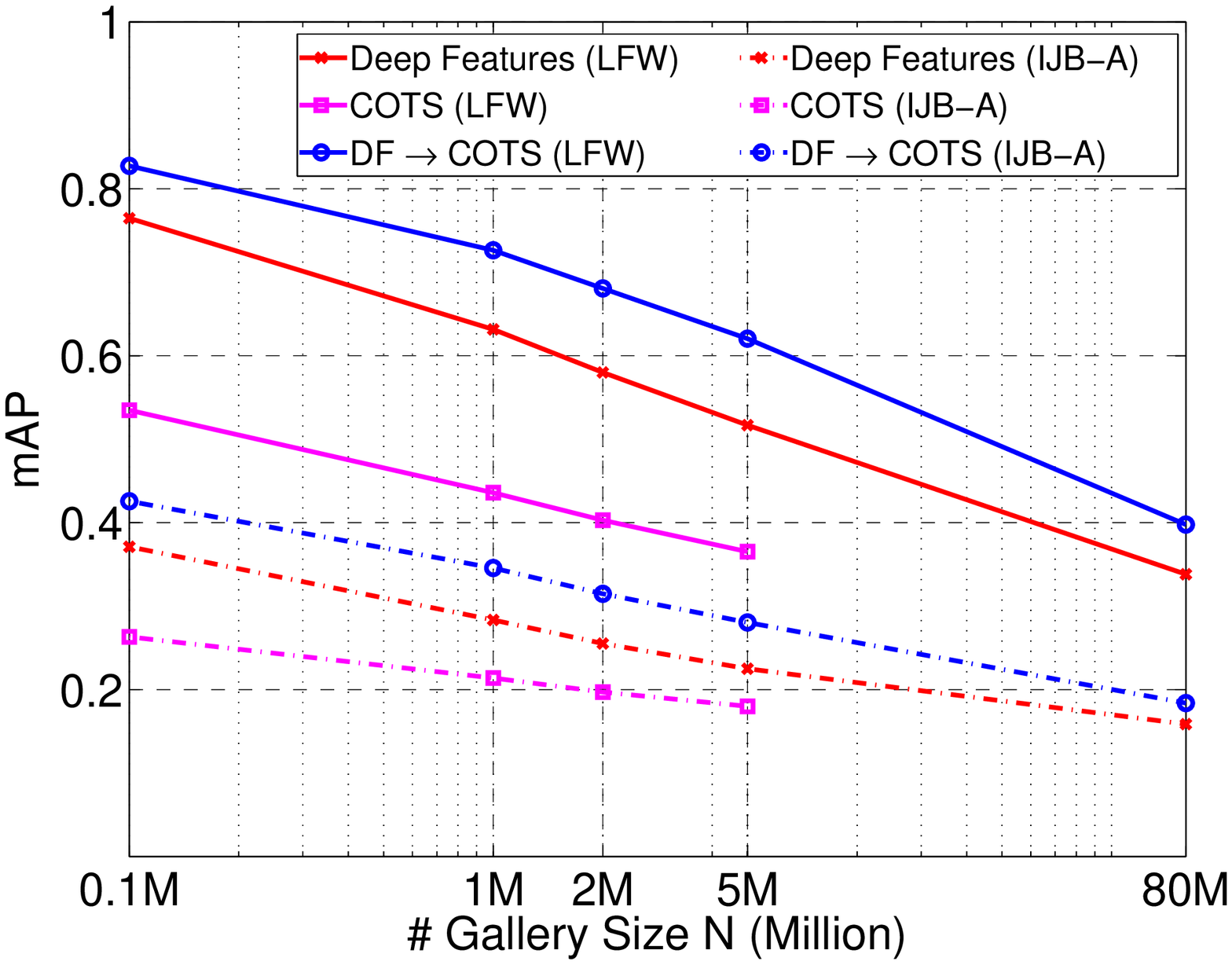}
  \makebox[3.4in]{Closed-set Search Evaluation on LFW and IJB-A datasets}
  \caption{Closed-set face search performance (mAP) vs. gallery size $N$ (log-scale), on LFW and IJB-A datasets. The performance of COTS matcher on $80$M gallery is not shown, since enrolling the complete $80$M gallery with the COTS matcher would have taken a prohibitive amount of time (over $80$ days).}
  \label{fig:exp_dbsize}
\end{figure}

\subsection{Closed-set Face Search}
We examine closed-set face search performance with varying gallery size $N$, from $100$K to $80$M. Enrolling the complete $80$M gallery in the COTS matcher would take a prohibitive amount of time (over $80$ days), due to limitations of the SDK we have, so the maximum gallery set used for the COTS matcher is $5$M. For the proposed face search scheme DF$\rightarrow$COTS, we chose the size of candidate set $k$ using the heuristic $k=1/100N$ when the gallery size is smaller than $5$M and $k=1,000$ when the gallery set size is $80$M. We use a fixed $k$ for the full $80$M gallery since using a  larger $k$ would take a prohibitive amount of time, due to the need to enroll the top-ranking images in the COTS matcher. Experimental results for the LFW and IJB-A datasets are shown in Figs.~\ref{fig:exp_dbsize}, respectively.

For both LFW and IJB-A face images, the recognition performance of all three face search schemes evaluated here decreases with increasing gallery set size. In particular, for all the search schemes, mAP linearly decreases with the gallery size $N$ on log scale; the performance gap between a $100$K gallery and a $5$M gallery is about the same as the performance gap between a $5$M gallery and an $80$M gallery. While deep features outperform the COTS matcher alone, the proposed cascaded face search system (which leverages both deep features and the COTS matcher) gives better search accuracy than either method individually. Results on the IJB-A dataset are overall similar to the LFW results. It is important to note that the overall performance on the IJB-A dataset is much lower than the LFW dataset, which is to be expected given the nature of the IJB-A dataset.
\begin{figure}[htbp]
  \centering
  \includegraphics[width=3.4in]{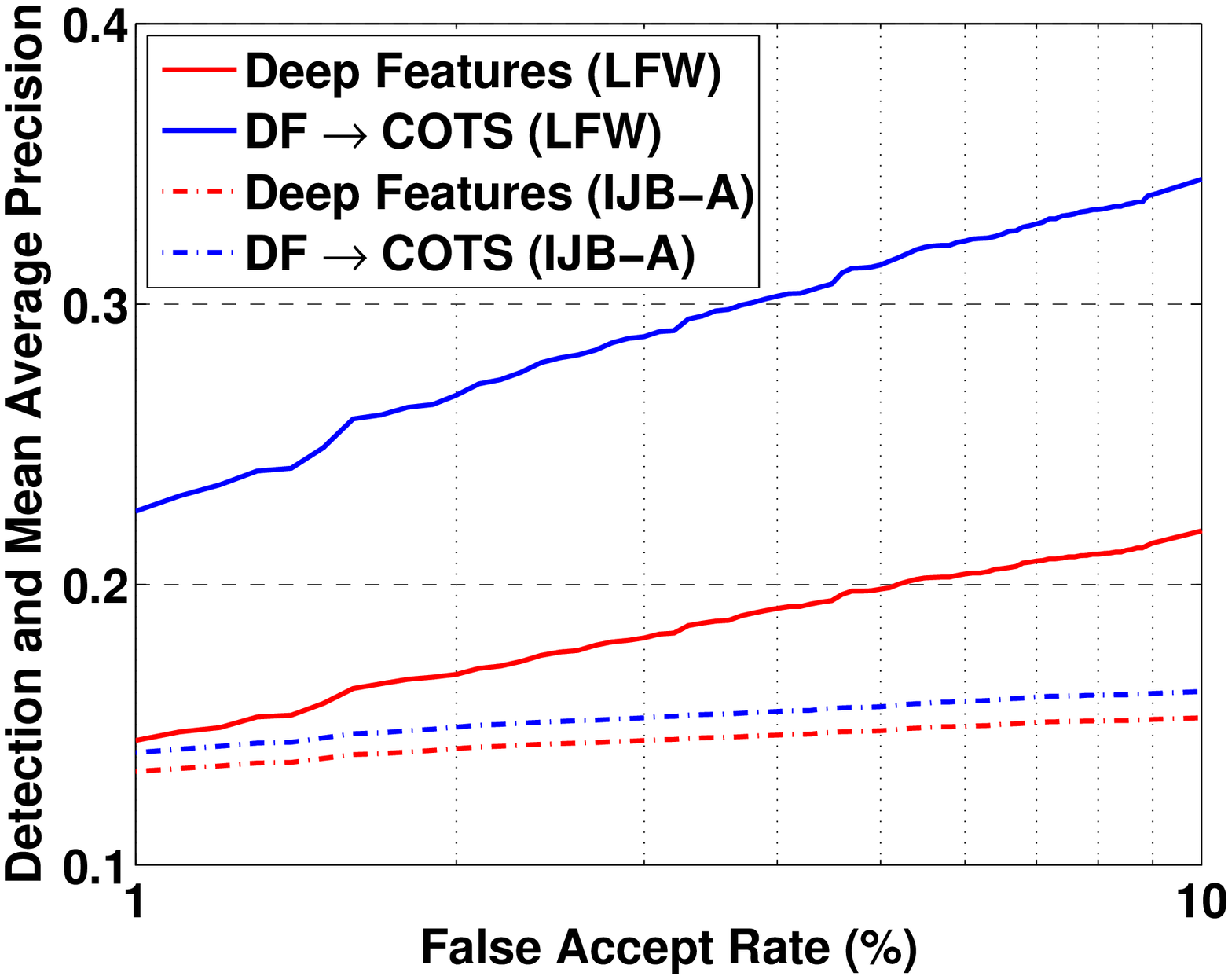}
  \makebox[3.4in]{Open-set Search Evaluation on LFW and IJB-A datasets}
  \caption{Open-set face search performance (mAP) vs. false accept rate (FAR) on LFW and IJB-A datasets, using an $80$M face gallery. The performance of COTS matcher is not shown, since enrolling the complete $80$M gallery with the COTS matcher would have taken a prohibitive amount of time (over $80$ days).}
  \label{fig:exp_openset}
\end{figure}

\subsection{Open-set Face Search}
Open-set search is important for several practical applications where it is unreasonable to assume that a gallery contains images of all potential probe subjects. We evaluate open-set search performance on an $80$M gallery, and plot the search performance (mAP) at varying FAR in Figs.~\ref{fig:exp_openset}.

For both the LFW and IJB-A datasets, the open-set face search problem is much harder than closed-set face search. At a FAR of $1\%$, the search performance (mAP) of the compared algorithms is much lower than the closed-set face search, indicating that a large number of genuine probe images are rejected at the threshold needed to attain $1\%$ FAR.

\subsection{Scalability}
In addition to mAP, we also report the search times in Table~\ref{tab:running_time}. We run all the experiments on a PC with an Intel(R) Xeon(R) CPU (E5-2687W) clocked at 3.10HZ. For a fair comparison, all the compared algorithms use only one CPU core. The deep features are extracted using a Tesla K40 graphics card.

In our experiments, template generation is applied over the entire gallery off-line, meaning that deep features are extracted for all gallery images and the gallery is indexed using product quantization before we begin processing probe images. We also enroll the gallery images using the COTS matcher and store the templates on disk. The run time of the proposed face search system after the gallery is enrolled and indexed consists of two parts: i) \emph{enrollment time} including face detection, alignment and feature extraction, and ii) \emph{search time} consisting of the time taken to find the top-$k$ search results given the probe template. Since we did not enroll all $80$M gallery images using the COTS matcher, we estimate the query time for the $80$M gallery by assuming that search time increases linearly with the gallery size.
\begin{table}[]
\scriptsize
\centering
\begin{threeparttable}[b]
\caption{The average search time (seconds) per probe face and the search performance (mAP).}\label{tab:running_time}
\begin{tabular}{l|ccc|ccc}
\toprule
               & \multicolumn{3}{c|}{5M Face Gallery} & \multicolumn{3}{c}{80M Face Gallery} \\ \cmidrule{2-7}
               & \multirow{2}{*}{COTS}   & \multirow{2}{*}{DF}   & {\scriptsize DF$\rightarrow$COTS} & \multirow{2}{*}{COTS}   & \multirow{2}{*}{DF}    & {\scriptsize DF$\rightarrow$COTS} \\
               & & & {\scriptsize @50K} & & & {\scriptsize @1K} \\ \midrule
Enrollment     & 0.09   & 0.05 & 0.14                & 0.09   & 0.05  & 0.14                \\
Search      & 30     & 0.84 & 1.15                & 480$^\star$    & 6.63  & 6.64                \\
Total Time     & 30.09  & 0.89 & 1.29                & 480.1$^\star$  & 6.68  & 6.88                \\ \cmidrule{1-1}
mAP            & 0.36   & 0.52 & 0.62                & N/A    & 0.34  & 0.4                 \\ \bottomrule
\end{tabular}
\begin{tablenotes}
    \item[$\star$]Estimated by assuming that search time increases linearly with gallery size.
\end{tablenotes}
\end{threeparttable}
\end{table}

Using product quantization for fast matching based on deep features, we can retrieve the top-$k$ candidate faces in about $0.9$ seconds for a $5$M image gallery and in about $6.7$ seconds for an $80$M gallery. On the other hand, the COTS matcher takes about $30$ and $480$ seconds to carry out brute-force comparison over the complete galleries of $5$ and $80$ million images, respectively. In the proposed cascaded face search system, we mitigate the impact of the slow exhaustive search required by the COTS matcher by only using them on a short candidate list. The proposed cascaded scheme takes about 1 second for the $5$M gallery and about $6.9$ seconds for the $80$M gallery, which is only a minor increase over the time taken using deep features alone ($6.68$ seconds). The search time could be further reduced by using a non-exhaustive search method, but that most likely will result in a significant loss in search accuracy.
\begin{figure}[htbp]
\centering
\includegraphics[width=3in]{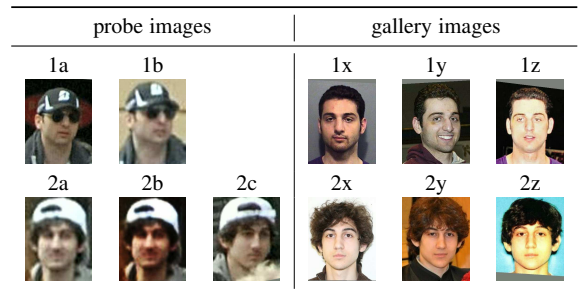}
\caption{Probe and gallery images of Dzhokhar Tsarnaev and Tamerlan Tsarnaev, responsible for the April 15, 2013 Boston marathon bombing. Face images 1a and 1b are the two probe images used for Suspect 1 (Dzhokhar Tsarnaev). Face images 2a, 2b and 2c are the three probe images used for Suspect 2 (Tamerlan Tsarnaev). The gallery images of the two suspects became available on media websites following the identification of the two suspects. Face images 1x, 1y and 1z are the three gallery images for Suspect 1 and images 2x, 2y and 2z are the three gallery images for Suspect 2.}
\label{fig:bombers}
\end{figure}

\begin{table*}[htbp]
\scriptsize
\centering
\caption{Rank search results of Boston bombers face search based on $5$M and $80$M gallery. The five probe images are designated as 1a, 1b, 2a, 2b, and 2c. The six mated images are designated as 1x, 1y, 1z, 2x, 2y, and 2z. The corresponding images are shown in Fig.~\ref{fig:bombers}}\label{tab:bomber}
\begin{tabular}{crrr|rrr|rrr}
\toprule
\multicolumn{1}{c|}{} & \multicolumn{3}{c|}{COTS (5M Gallery)} & \multicolumn{3}{c|}{Deep Features (5M Gallery)} & \multicolumn{3}{c}{Deep Features (80M Gallery)}\\
\cmidrule{2-10}
\multicolumn{1}{c|}{} & 1x & 1y & 1z & 1x & 1y & 1z & 1x & 1y & 1z \\
\cmidrule{1-10}
\multicolumn{1}{c|}{1a}  & 2,041,004     & 595,265       & 1,750,309     & 132,577       & 232,275       & 1,401,474                & 2,566,917  & 5,398,454     & 31,960,091    \\
\multicolumn{1}{c|}{1b}  & 3,816,874     & 3,688,368     & 2,756,641     & 1,511,300     & 1,152,484     & 1,699,926                & 33,783,360 & 27,439,526    & 44,282,173    \\
\cmidrule{1-10}
\multicolumn{1}{c|}{} & 2x & 2y & 2z & 2x & 2y & 2z & 2x & 2y & 2z \\
\cmidrule{1-10}
\multicolumn{1}{c|}{2a}& 67,766        & 86,747        & 301,868       & 174,438       & 39,417        & 105879                 & 2,461,664  & 875,168       & 1,547,895     \\
\multicolumn{1}{c|}{2b} & 352,062       & 48,335        & 865,043       & 71,783        & 26,525        & 84,012                  & 1,417,768  & 972,411       & 1,367,694     \\
\multicolumn{1}{c|}{2c} & 158,341       & 625           & 515,851       & {\bf 9}       & {\bf 341}     & 9,975                   & {\bf 109}  & 2,952         & 136,651       \\
\cmidrule{1-10}
& Proposed Cascaded Face Search System\\
\cmidrule{1-10}
\multicolumn{1}{c|}{2c} & \multicolumn{3}{l|}{DF$\rightarrow$COTS@1K}   & {\bf  7}      & {\bf 1}       & 9,975 & {\bf 46}   & 2,952         & 136,651 \\
\multicolumn{1}{c|}{2c} & \multicolumn{3}{l|}{DF$\rightarrow$COTS@10K}  & {\bf 10}      & {\bf 1}       & 1,580 & {\bf 160}  & {\bf 8}       & 136,651 \\
\bottomrule
\end{tabular}
\end{table*}

\section{Boston Marathon Bombing Case Study}
In addition to the large-scale face search experiments reported above, we report on a case-study: finding the identity of Boston marathon bombing suspects\footnote{\url{https://en.wikipedia.org/wiki/Boston_Marathon_bombing}} in an $80$M face gallery.

Klontz and Jain~\cite{faceretrieval:bomber} made an attempt to identify the face images of the Boston marathon bombing suspects in a large gallery of mugshot images. Video frames of the two suspects were matched against a background set of mugshots using two state-of-the-art COTS face matcher. Five low resolution images of the two suspects, released by the FBI (shown in left side of Fig.~\ref{fig:bombers}) were used as probe images, and six images of the suspects released by the media (shown in the right side of Fig.~\ref{fig:bombers}) were used as the mates in the gallery. These gallery images were augmented with $1$ million mugshot images. One of the COTS matchers was successful in finding the true mate (2y) of one of  the probe image (2c) of Tamerlan Tsarnaev at rank $1$.

To evaluate the face search performance of our cascaded face search system, we construct a similar search problem under more challenging conditions by adding the six gallery images to a background set of up to $80$ million web faces. We argue that the unconstrained web faces are more consistent with the quality of the images of the suspects used in the gallery than mugshot images and therefore comprise a more meaningful gallery set. We evaluate the search results using gallery sizes of 5M and $80$M leveraging the same background set used in our prior search experiments. Since there is no demographic information available for the web face images we downloaded, we only conduct a ``blind search"~\cite{faceretrieval:bomber}, and do not filter the gallery using any demographic information.

The search results are shown in Table~\ref{tab:bomber}. Both the deep features and the COTS matcher fail on probe images 1a, 2b, 2a, and 2b, similar to the results in~\cite{faceretrieval:bomber}. On the other hand, for probe 2c, the deep features perform much better than the COTS matcher. For the $5$M gallery, the COTS matcher found a mate for probe 2c at rank $625$; however, the deep features returned the gallery image 2x at rank $9$. The proposed cascaded search system returned gallery image 2y at rank $1$ in the 5M image gallery, by using the COTS matcher to re-rank the deep features results, demonstrating the value of the proposed cascade framework. Results are somewhat worse for the $80$M image gallery. For probe 2c, using deep features alone, we find gallery image 2x at rank $109$ and gallery image 2y at rank $2,952$. However, using the cascaded search system, we find gallery image 2x at rank $46$ by re-ranking the top-$1$K faces, and find gallery image 2y at rank $8$ by re-ranking the top-$10$K faces. So, even with an $80$M image gallery, we can successfully find a match for one of the probe image (2c) within the top-10 retrieved faces.

\begin{figure*}[htbp]
    \centering
    \includegraphics[width=6.6in]{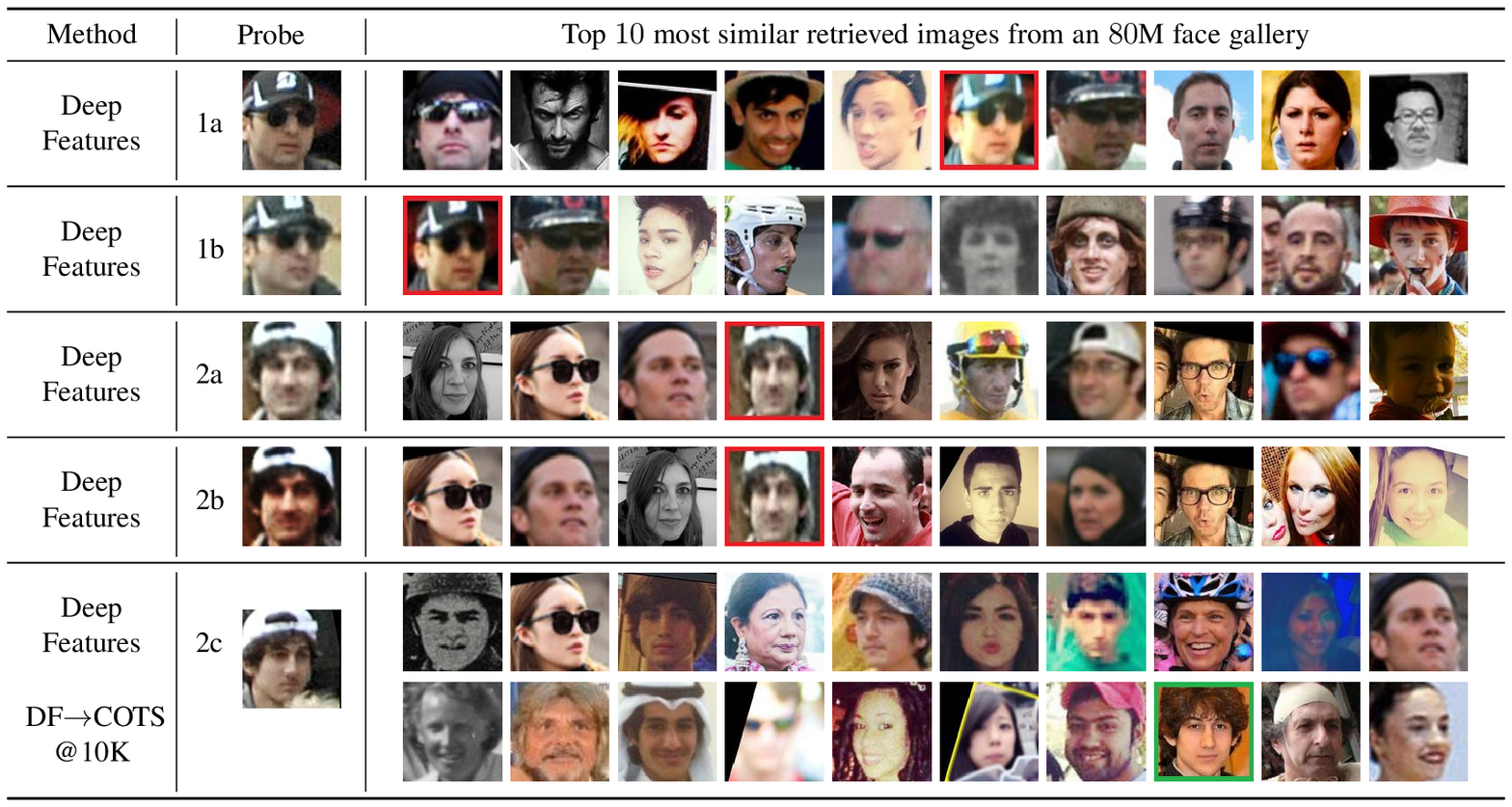}\vspace{-0.1in}
    \caption{Top $10$ search results for the two Boston marathon bombers on the $80$M face gallery. The first two probe faces  are of the older brother (Dzhokhar Tsarnaev) and the last three probe faces are of the younger brother (Tamerlan Tsarnaev). For each probe face, the retrieved image with \textcolor{green}{green} border is the correctly retrieved image. Images with the \textcolor{red}{red} border are near-duplicate images present in the gallery. Note that we were not aware of the existence of these near-duplicate images in the gallery before the search.} \label{fig:visual_bomber_80}
\end{figure*}

The face search results for the $80$M galleries are shown in Fig.~\ref{fig:visual_bomber_80}. One interesting observation is that deep features will typically return faces taken under similar conditions to the probe image. For example, a list of candidate images with sunglasses are returned for probe image, which exhibits partial occlusion due to sunglasses. Similarly, a list of blurred candidate faces are returned for probe, which is of low resolution due to blur. Another interesting observation is that the deep features based search found several near-duplicate images which happened to be present in the unlabeled background dataset, images which we were not aware of prior to viewing these search results.

\newcommand{\includeBomberVisual}[1]{\includegraphics[width=0.45in]{#1}}

\section{Conclusions}
We have proposed a cascaded face search system suitable for large-scale search problems. We have developed a deep learning based face representation trained on the publicly available CASIA dataset~\cite{DB:CASIA}. The deep features are used in a product quantization based approximate $k$-NN search to first obtain a short list of candidate faces. This short list of candidate faces is then re-ranked using the similarity scores provided by a state-of-the-art COTS face matcher. We demonstrate the performance of our deep features on three face recognition datasets, of increasing difficulty: the PCSO mugshot dataset, the LFW unconstrained face dataset, and the IJB-A dataset. On the mugshot data, our performance (TAR of $93.5\%$ at FAR of $0.01\%$) is worse than a COTS matcher ($98.5\%$), but fusing our deep features with the COTS matcher still improves overall performance ($99.2\%$). Our performance on the standard LFW protocol ($98.23\%$ accuracy) is comparable to state of the art accuracies reported in the literature. On the BLUFR protocol for the LFW database we attain the best reported performance to date (verification rate of $87.65\%$ at FAR of $0.1\%$). We outperform the benchmarks reported in~\cite{db:janus} on the IJB-A dataset, as follows: TAR of $51.4\%$ at FAR of $0.1\%$ (verification); Rank 1 retrieval of $82.0\%$ (closed-set search); FNIR of $61.7\%$ at FPIR of $1\%$ (open-set search). In addition to the evaluations on the LFW and the IJB-A benchmarks, we evaluate the proposed search scheme on an 80 million face gallery, and show that the proposed scheme offers an attractive balance between recognition accuracy and runtime. We also demonstrate search performance on an operational case study involving the video frames of the two persons (Tsarnaev brothers) implicated in the 2013 Boston marathon bombing. In this case study, the proposed system can find one of the suspects' images at rank 1 in 1 second on a 5M gallery and at rank 8 in 7 seconds on an 80M gallery.

We consider non-exhaustive face search an avenue for further research. Although we made an attempt to employ indexing methods, they resulted in a drastic decrease in search performance. If only a few searches need to be made, the current system's search speed is adequate, but if the number of searches required is on the order of the gallery size, the current runtime is inadequate. We are also interested in improving the underlying face representation, via improved network architectures, as well as larger training sets.

\bibliographystyle{IEEEtran}
\balance

\end{document}